\pdfoutput=1

\documentclass[11pt]{article}

\usepackage[]{acl}

\usepackage{times}
\usepackage{latexsym}

\usepackage[T1]{fontenc}

\usepackage[utf8]{inputenc}

\usepackage{microtype}

\usepackage{inconsolata}

\usepackage{graphicx}

\usepackage{float}

\usepackage{booktabs}
\usepackage{adjustbox}
\usepackage{textcomp}
\usepackage{multirow}
\usepackage{xcolor}
\usepackage{colortbl}
\usepackage{todonotes}
\usepackage{enumitem}
\usepackage{footmisc}
\usepackage{amsmath}
\usepackage{comment}

\definecolor{ao(english)}{rgb}{0.0, 0.5, 0.0}
\definecolor{cornellred}{rgb}{0.7, 0.11, 0.11}
\definecolor{pf7}{RGB}{166, 118, 29}

\newcommand{\furl}[1]{\footnote{\url{#1}}}

\usepackage{pgfplots}
\usepackage{filecontents}
\usepackage{subcaption} 

\newcolumntype{R}[2]{%
    >{\adjustbox{angle=#1,lap=\width-(#2)}\bgroup}%
     l%
    <{\egroup}%
}

\title{Balancing Continuous Pre-Training and Instruction Fine-Tuning: Optimizing Instruction-Following in LLMs}

\author{
  \textbf{Ishan Jindal},
  \textbf{Chandana Badrinath},
  \textbf{Pranjal Bharti}, \\
  \textbf{Lakkidi Vinay}, 
  \textbf{Sachin Dev Sharma}
\\
  Samsung Research\\
 {\tt \{ishan.jindal, c.badrinath, p.bharti, l.vinay, sachin.dev\}@samsung.com}
}

\begin{document}
\maketitle
\begin{abstract}
Large Language Models (LLMs) for public use require continuous pre-training to remain up-to-date with the latest data. The models also need to be fine-tuned with specific instructions to maintain their ability to follow instructions accurately. Typically, LLMs are released in two versions: the Base LLM, pre-trained on diverse data, and the instruction-refined LLM, additionally trained with specific instructions for better instruction following. The question arises as to which model should undergo continuous pre-training to maintain its instruction-following abilities while also staying current with the latest data. In this study, we delve into the intricate \textbf{relationship between continuous pre-training and instruction fine-tuning} of the LLMs and investigate the impact of continuous pre-training on the instruction following abilities of both the base and its instruction finetuned model. Further, the instruction fine-tuning process is computationally intense and requires a substantial number of hand-annotated examples for the model to learn effectively. This study aims to find the most \textbf{compute-efficient strategy} to gain up-to-date knowledge and instruction-following capabilities without requiring any instruction data and fine-tuning. We empirically prove our findings on the LLaMa 3, 3.1 and Qwen 2, 2.5 family of base and instruction models, providing a comprehensive exploration of our hypotheses across varying sizes of pre-training data corpus and different LLMs settings.

\end{abstract}

\section{Introduction}

Recently, autoregressive large language models (LLM) showed remarkable progress across a wide range of natural language tasks, natural language understanding, mathematical reasoning, and coding across various domains \cite{achiam2023gpt,team2024gemma,touvron2023llama,roziere2023code,yang2024qwen2}. These LLMs are pre-trained with a causal language modeling objective to predict the next token(s) in a given sequence until it is complete, termed as \emph{Base models}. These base models exhibit a remarkable ability to generate linguistically coherent text, however not necessarily aligning their generations with human preferences and needs \cite{ouyang2022training}. Thus, LLMs often require a fine-tuning step, \emph{Instruction fine-tuning} to bridge the gap between the base model's fundamental objective and the practical needs of human users \cite{rafailov2024direct,ethayarajh2024kto} termed as \emph{Instruction models}.

Instruction fine-tuning is an expensive task and generally requires a significant amount of labeled data\footnote{10M hand-annotated examples were used to instruction fine-tune LLaMa 3 instruct model \cite{llama3modelcard}.} depending on the type of optimization technique used\footnote{refer to Appendix \ref{app:flops} for more insights}. This can be expensive and time-consuming to collect and annotate such a big dataset. Algorithmically, it requires training of reward model and RLHF, PPO \cite{ouyang2022training}, DPO \cite{rafailov2024direct} fine-tuning which further adds to the complexity of the task. 

Parallelly, to stay abreast with the latest data, the base model needs to be either re-pre-trained on a combination of old and newly collected data \cite{gao2020pile,tokpanov2024zyda} or continuously pre-trained on the newly collected data \cite{ibrahim2024simple} yielding to the new base model.  For example, the LLaMa 3.1 base model is pre-trained with more and high-quality data over the LLaMa 3 base model \cite{dubey2024llama}. Similarly, Qwen 2.5 family base models have more knowledge and improved capabilities over Qwen 2 family models \cite{qwen2.5}.  

Continuous pre-training of the LLM generally results in forgetting previously learned information, several methods have been proposed to maintain the base model performance on previously learned tasks such as \citet{xie2023efficient,ibrahim2024simple}. However, there has been no research focusing on the influence of continuous training on instruction models. As continuous pre-training is vital for acquiring new knowledge, and instruction tuning is necessary to learn instruction following capabilities, it is required to have both the capabilities to any instruction model.  This raises a series of natural questions:

\begin{itemize}
\item[a] What happens to the instruction capabilities when we continuously pre-train the instruction model to gain new knowledge?
\item[b] If lost, how to regain instruction capabilities? 
\item[c] Is it necessary to add resource-extensive instruction-fine-tuning after updating the knowledge of the base model?
\end{itemize}

We approach this problem empirically by studying two different settings. In the first setting, we continuously pre-train the instruction model on a specific dataset and observe its performance on the LLM harness framework from EleutherAI \cite{gao2021framework}. Whereas in another setting we continuously pre-train the base model with the same data and then instruction fine-tune the continuously pre-trained base model. Finally, we compare the instruction capabilities of instruction models from both settings. Since instruction fine-tuning is an expensive task, we discovered a simple yet efficient approach to regain the instruction capability of the continuous pre-trained base model, given that the instruction-tuned model of the original base model is available. 
Our main findings and the contributions of this work are as follows:
\begin{itemize}
  \setlength{\parskip}{0pt}
  \setlength{\itemsep}{0pt plus 1pt}
\item Continuous pre-training of an instruction model results in catastrophic forgetting of the instruction capabilities and, therefore should be avoided. Section \ref{ssec:icp}.
\item Continuous pre-training base model and then instruction tuning preserve both the domain knowledge and the instruction capabilities. Section \ref{ssec:ira}
\item Instruction capabilities are portable across the same ancestor models. That is, we can extract the instruction capability by simply subtracting the weight of the base model from the weights of its instruction-tuned model. Section \ref{ssec:ipl}
\item No traditional instruction tuning is required for a continuous pre-trained base model instead the instruction capabilities are ported. Section \ref{ssec:RIC}
\end{itemize}

To our knowledge, we are the first ones to systematically conduct this analysis and discover the portability of the instruction capabilities across models from the same ancestor. We empirically prove all our findings on LLaMa 3, LLaMa 3.1, Qwen2, and Qwen 2.5 families of base and instruct models. We comprehensively test our hypothesis in breadth and depth with varying sizes of pre-training data corpus across different LLMs settings in Section \ref{sec:exp}.

\section{Background}
\label{sec:ir}
In this investigation, we focus on the LLM families for which both the base model and the corresponding instruction-tuned model are publicly available. Let $\theta^{d1}_{b}$ be the learned parameters of the autoregressive base model on some pre-trained dataset $d1$, and $\theta^{d1v1}_{i}$ the corresponding parameters of instruction tuned LLM  fine-tuned on some instruction dataset $v1$. Here, instruction tuning is applied on top of the base model. Given a new pre-training dataset $d2$ our objective is to obtain a $d2$ specific LLM (resulted LLM) that has the following two properties
\begin{itemize}
  \setlength{\parskip}{0pt}
  \setlength{\itemsep}{0pt plus 1pt}
\item[P1] Since the $d2$ is not significantly large (as compared to $d1$) we do not want resulted LLM to forget the language understanding capabilities of the base model that it learned during the very first iteration of pre-training. Here, $d2$ itself is not sufficiently large ($<$1B tokens) to bring language understanding capabilities to any moderate-scale LLM (say 7B).
\item[P2] Further, to align the resulted model generations with human needs, the resulted model should also have instruction following capabilities at least of the same levels as the base model.
\end{itemize} 
There could be two possible settings to bring both the above properties to any LLM. 

\begin{enumerate}
\item[S1] Directly start with the instruction tuned LLM $\theta^{d1v1}_{i}$ and continuous pre-train with $d2$ dataset assuming that the resulted LLM will possess the above two properties P1, P2. 
\item[S2] First, continuously pre-train the $\theta^{d1}_{b}$ on $d2$ pre-training dataset  via continual pre-training from \citet{ibrahim2024simple} avoiding catastrophic forgetting of $d1$ learned knowledge (gain P1). Then instruction was fine-tuned with $v1$ dataset to gain instruction following capabilities (gain P2).

\end{enumerate} 

\subsection{Resulted LLM}
In this section, we explore both the above settings analyzing the anticipated pros and cons.
\subsubsection{Setting 1: Continuous Pre-training of Instruction-Tuned LLM }
With an assumption that the instruction capabilities of the instruction-tuned LLM will not get lost with continuous pretraining on some raw data, $d2$ is the least expensive setting to get both the new knowledge and the instruction capabilities. However,  in our experiments, we could not find any evidence to validate this assumption instead observed the null hypothesis. Having said so, let's denote $\theta^{d1v1d2}_{i}$ are the parameters of the $d2$ continuously pre-trained instruction-tuned LLM. 

\subsubsection{Setting 2: Continuous Pre-training of Base LLM followed by Instruction Fine-tuning}
In this setting, the base model (parameters $\theta^{d1}_{b}$) is first continuously pre-trained on $d2$ dataset resulting in a new base model with parameters $\theta^{d1d2}_{b}$. This updated base model has learned the new domain-specific knowledge without forgetting the initially learned knowledge \cite{ibrahim2024simple}. Now, to add the instruction following capabilities this new base model needs to be instruction tuned. Let's denote $\theta^{d1d2v1}_{i}$ are the parameters of the resulting LLM that is instruction tuned on $d2$ continuously pre-trained LLM. 

To perform instruction tuning, first, high-quality instruction-formatted data needs to be collected or constructed. Then, these formatted instances are used to fine-tune LLM in a supervised learning fashion. However, instruction tuning is an expensive time-consuming task and often poses many challenges such as clean instruction-formatted dataset \cite{sun2024principle},  instruction task formatting and design \cite{wang2022super}, instruction optimization and its scalability \cite{xu2023wizardlm}, and other practical issues with instruction fine-tuning. With the availability of the original instruction fine-tuning dataset\footnote{Instruction fine-tuning datasets used to tune LLaMa, Qwen, and other SoTA LLMs are not shared in public} some of the above-mentioned challenges might be resolved however, the practical issues such as fine-tuning stability remains.

\subsection{Instruction Residuals}
In this section, we describe the instruction residual approach to simply regain the instruction following capabilities \footnote{Instruction following capabilities and instruction capabilities are used interchangeably in this work}. We compute the instruction residual between a instruction following LLM $\theta^{d1v1}_{i}$ and its corresponding base model $\theta^{d1}_{b}$ in the parametric space as 
\begin{equation}
\Theta^{v1}_{r} = \theta^{d1v1}_{i} - \theta^{d1}_{b}.
\label{eq:ir}
\end{equation}

This residual computation is inspired by the parameter efficient fine-tuning of LLMs such as low-rank adaptation (LoRA \cite{hulora}, QLoRA\cite{dettmers2024qlora}, DoRA \cite{liu2024dora} etc). In these techniques instead of fine-tuning a large weight matrix $W$ for a given layer, a low-rank $\Delta W$ matrix is learned, which contains the new information to be integrated with the original model that is $W_{updated} = W + \Delta W$. These techniques add new information/capabilities to the original model often with fewer parameters defined by \emph{rank} of  $\Delta W$. With the full $\Delta W$ rank, it is similar to fine-tuning the whole model end-to-end \cite{hulora}.

Inspired by this idea of weight addition to learning a new capability, we first extract the instruction capability by subtracting the base LLM weights from its corresponding instruction-tuned LLM weights as in \ref{eq:ir}, termed as \emph{instruction residuals}, and add this instruction residual to the continuously pre-trained base LLM on new skill $d2$ that is 

\begin{equation}
\theta^{d1d2v1}_{i} = \theta^{d1d2}_{b}  \oplus \Theta^{v1}_{r}, 
\label{eq:update}
\end{equation}

where $\oplus$ represents element-wise addition. These tensor addition and subtraction to regain the instruction capabilities do not incur heavy computation costs making the instruction-tuned LLM readily available once the new knowledge is learned by the base LLM. One major limitation of this work is that if the base LLM and its corresponding instruction-tuned LLM are not available then the instruction residuals from \ref{eq:ir} won't be available and hence requires a full cycle of instruction fine-tuning to regain this ability. 

\section{Experiments}
\label{sec:exp}
\subsection{Datasets}
\subsubsection{Pre-traininig Dataset} 
\label{ssec:pretrain}
We need this pre-training dataset to test the impact on instruction capabilities of the continuously pre-trained model. We want the new pre-training data such that none of the base models and the corresponding instruct model have seen that data previously. Since the data contamination is a serious concern as noted in \citet{jiang2024does}, the existing pre-training datasets may not be the right choice to continuously pre-train the model. Therefore,  we manually scraped around 2M articles using a static news crawler FUNDUS\footnote{\url{https://github.com/flairNLP/fundus}} \cite{dallabetta-etal-2024-fundus}. We choose the news articles that are new to LLaMa 3.1 models that is we choose the articles published in the date range from December 2023 to September 2024 from all existing publishers in FUNDUS. The average length of the articles is 650 LLaMa tokens with 6981 max tokens and 156 min tokens. These articles are then packed with a sequence length of 4096 (LLaMa maximum sequence length is 8K but we choose 4K to efficiently utilize the existing GPU vRAM), and similar to \citet{kosec2021packing} we use attention masks for each article to avoid cross article contamination.

\subsubsection{Evaluation Dataset}
\begin{table}[!t]
\centering
\begin{adjustbox}{width=0.95\linewidth}
    \begin{tabular}{ l l l  }
    \toprule
        Category & Sub Category & Benchmark  \\ \midrule
        \multirow{4}{*}{\begin{tabular}[c]{@{}c@{}}Instruction\\ following\end{tabular}} & \multirow{3}{*}{\begin{tabular}[c]{@{}c@{}}Language\\ understanding\end{tabular}}  & IFEval  \\ \cmidrule{3-3}
        & & MMLU  \\ 
        &  &MMLU-Pro  \\ \cmidrule{2-3}
        & Math and logic & GSM8K  \\ \midrule
          \multirow{4}{*}{\begin{tabular}[c]{@{}c@{}}Reasoning\\ and \\  problem \\ solving\end{tabular}}  &\multirow{2}{*}{Commonsense} & Winogrande  \\ 
        &  & Hellaswag  \\ \cmidrule{2-3}
        & Factual knowledge & ARC\_easy  \\ \cmidrule{2-3}
        & Physical reasoning & Piqa  \\ \midrule
        Truthfulness & ~ & Truthfulqa\_mc2  \\ \bottomrule
    \end{tabular}
\end{adjustbox}
\caption{Evaluation dataset categorization.}
\label{tab:eval_data}
\end{table}

In this section, we describe the test dataset used to evaluate our hypothesis. To perform a comprehensive evaluation and to maintain reproducibility we use the evaluation harness framework from EleutherAI \cite{gao2021framework}. We particularly target to evaluate the following capabilities: 

\noindent\textbf{Instruction following}

\noindent\textbf{IFEval} This dataset was introduced mainly to focus on natural language instruction following capabilities of LLMs \cite{zhou2023instruction}. It contains 25 types of verifiable instructions such as  \textit{write in more than 400 words}, \textit{mention the keyword of AI at least 3 times } with 500 prompts. This evaluation is performed on  4 metrics: (1) Prompt-level strict-accuracy (PLS-acc): The percentage of prompts that all verifiable instructions in each prompt are followed, (2) Inst-level strict-accuracy (ILS-acc): The percentage of verifiable instructions that are followed, (3) Prompt-level loose-accuracy (PLL-acc): Prompt-level accuracy computed with the loose criterion, and (4) Inst-level loose-accuracy (ILL-acc): Instruction-level accuracy computed with a loose criterion.

\noindent\textbf{MMLU} mainly focuses on extensive world knowledge across 57 subjects which includes all major domains like math, computer science, medicine, philosophy, and law \cite{hendrycks2020measuring}. This dataset contains a total of 15908 development and test questions with 4 possible answers each. 

\noindent\textbf{MMLU-Pro} is introduced to further increase the complexity of the MMLU benchmark since the existing LLMs are excelled at MMLU \cite{wang2024mmlu}. This dataset was curated by eliminating some trivial and noisy questions from MMLU and by introducing reasoning-focused questions to MMLU which has mostly knowledge-driven questions.

\noindent\textbf{GSM8K} dataset consists of 8.5K high-quality linguistically diverse grade school math problems \cite{cobbe2021training}. These problems take between 2 and 8 steps to solve, and solutions primarily involve performing a sequence of elementary calculations using basic arithmetic operations ($+, -, \times, \div $) to reach the final answer. 
\noindent\textbf{Reasoning and problem-solving}

\noindent\textbf{Winogrande} is a large scale 44k commonsense reasoning dataset. It mainly tests a model's ability to resolve ambiguous pronouns based on contextual understanding \cite{sakaguchi2021winogrande}.

\noindent\textbf{Hellaswag}  is designed to benchmark commonsense reasoning in AI models \cite{zellers2019hellaswag}. It contains 10,000 multiple-choice questions for validation and testing. The dataset focuses on predicting the most plausible continuation of a given scenario.

\begin{table*}[!t]
    \centering
\begin{adjustbox}{width=0.97\linewidth}
\begin{tabular}{ll| llll | llll | lll}
\toprule
Benchmark                   &  \multicolumn{1}{c}{Metric}   & \multicolumn{4}{c}{L3b} & \multicolumn{4}{c}{L3i} & \multicolumn{3}{c}{L3b + 3Lr} \\ \cmidrule(lr){3-6} \cmidrule(lr){7-10} \cmidrule(lr){11-13}
\multicolumn{2}{c}{\#   of new tokens $\rightarrow$ } & org.      & +100M  & +500M   &  \multicolumn{1}{l}{+1B}    & org.     & +100M  & +500M   &  \multicolumn{1}{l}{+1B}    & 100M & 500M          & 1B           \\ \midrule
\multirow{4}{*}{IFEval}     & ILL\_acc & 19.06 & 17.75 & 19.30 & 20.14 & 53.36 & 45.68 & 45.32 & 41.01 & 57.67 & 56.47 & 57.79 \\
                            & ILS\_acc & 17.87 & 16.31 & 17.87 & 17.87 & 47.84 & 40.41 & 38.61 & 35.25 & 51.68 & 51.44 & 51.68 \\
                            & PLL\_acc & 09.98  & 09.61  & 10.54 & 11.09 & 41.77 & 34.20 & 33.64 & 28.10 & 44.18 & 42.88 & 44.36 \\
                            & PLS\_acc & 09.06  & 08.69  & 09.61  & 09.80  & 35.30 & 29.21 & 26.80 & 22.55 & 37.52 & 36.78 & 37.52 \\ \cmidrule{2-13}
MMLU                        & acc      & 62.14 & 63.76 & 63.77 & 63.62 & 63.83 & 66.04 & 65.38 & 65.52 & 67.69 & 67.16 & 67.51 \\
MMLU-Pro                    & EM       & 34.51 & 34.92 & 34.70 & 35.49 & 39.70 & 36.39  & 35.76 & 35.52 & 40.72  & 40.84 & 40.27 \\ \cmidrule{2-13}
\multirow{2}{*}{GSM8K}      & EM       & 49.58 & 48.14 & 47.54 & 47.08 & 75.06 & 69.22 & 68.08 & 68.01 & 74.83 & 73.92 & 73.16 \\
                            & strict-EM       &49.20 & 34.04 & 36.62 & 39.04 & 74.98 & 65.35 & 59.74 & 55.42 & 46.93 & 55.27 & 49.51 \\  \midrule \midrule
\multicolumn{2}{c||}{Sub-Average} & 31.43 & 29.15 & 29.99 & 30.52 & \textbf{53.98} & 48.31 & 46.67 & 43.92 & 52.65 & \underline{53.10} & 52.73 \\ \midrule \midrule 
Winogrande                  & acc      &73.16 & 72.14 & 71.59 & 71.27 & 71.74 & 72.22 & 71.74 & 71.74 & 71.43 & 72.06 & 71.19 \\ \cmidrule{2-13}
\multirow{2}{*}{Hellaswag}  & acc      & 60.12 & 60.17 & 60.27 & 60.23 & 57.70 & 58.69 & 58.73 & 59.00 & 59.30 & 59.16 & 59.39 \\
                            & acc\_n   & 79.22 & 78.62 & 78.20 & 78.36 & 75.76 & 77.84 & 77.68 & 77.80 & 79.01 & 79.14 & 79.00 \\ \cmidrule{2-13}
\multirow{2}{*}{ARC\_easy}  & acc      & 80.39 & 81.14 & 80.60 & 80.60 & 81.52 & 79.71 & 79.63 & 79.46 & 82.20 & 82.32 & 82.03 \\
                            & acc\_n   &77.78 & 80.30 & 79.67 & 79.38 & 79.63 & 77.53 & 77.65 & 77.44 & 79.00 & 79.25 & 79.08 \\ \cmidrule{2-13}
\multirow{2}{*}{Piqa}       & acc      & 79.54 & 80.09 & 80.30 & 80.20 & 78.56 & 79.87 & 79.49 & 79.49 & 80.25 & 80.20 & 80.41 \\
                            & acc\_n   & 80.74 & 81.56 & 81.34 & 80.96 & 78.62 & 80.41 & 79.98 & 79.98 & 80.63 & 80.74 & 80.90 \\  \midrule \midrule
\multicolumn{2}{c||}{Sub-Average} & 75.85 & \textbf{76.29} & 76.00 & 75.86 & 74.79 & 75.18 & 74.99 & 74.99 & 75.97 & \underline{76.12} & 76.00 \\ \midrule \midrule 
T\_mc2             & acc      & 43.94 & 47.64 & 47.29 & 47.41 & 51.67 & 51.80 & 51.55 & 51.68 & \underline{56.13} & 55.73 & \textbf{56.17} \\ \midrule
\multicolumn{2}{c|}{Average}         &51.64 & 50.93 & 51.20 & 51.41 & 62.94 & 60.28 & 59.36 & 58.00 & 63.07 & 63.34 & 63.12
 \\ \bottomrule

\end{tabular}
\end{adjustbox}
\caption{Impact of continual pretraining on LLaMa 3 Base (L3b) and the LLaMa 3 instruction tuned (L3i) models w.r.t varying number of new tokens. Also, depicts the usefulness of the instruction residual technique to regain instructional capabilities.}
\label{tab:results_pretrain}
\end{table*}

\noindent\textbf{ARC\_easy} dataset consists of a collection of 7787 natural science questions \cite{clark2018think}. The dataset contains only natural, grade-school science questions. ARC questions appeal to both different styles of knowledge and different styles of reasoning.

\noindent\textbf{Piqa} evaluates the model on the physical commonsense questions without experiencing the physical world \cite{bisk2020piqa}. Each instruction has a goal to reach in the physical world, given the description of the environment if required, and 2 options (solutions) to reach the goal. We report both the accuracy and the normalized accuracy on this dataset.

\noindent\textbf{Truthfulness} measure the truthfulness of a language model in answering questions \cite{lin2021truthfulqa}. This dataset consists of 817 questions across 38 categories and captures human misconceptions, false beliefs, conspiracies, and awareness between real-world knowledge and fictional knowledge across 38 domains including health, law, finance, and politics. Because of space constraints, we abbreviate this dataset as \emph{T\_mc2}.

We choose these datasets as these are commonly evaluated for most of the newly released LLMs \cite{touvron2023llama,yang2024qwen2}.  Table \ref{tab:eval_data} summarizes all the evaluation datasets used in this work for each category.  We used the latest versions of these datasets available on EleutherAI\footnote{\url{https://github.com/EleutherAI/lm-evaluation-harness}} as of writing this work. Only MMLU, MMLU-Pro, and GSM8K are evaluated with 5-shot, rest of the datasets are evaluated on zero-shot.

\begin{figure*}[t!]
    \centering
    \begin{subfigure}[t]{0.32\textwidth}
        \begin{tikzpicture}
            \begin{axis}[
                ybar, 
                ymajorgrids=true,
                bar width=6.5pt,
                width=180pt,
                height=6.5cm,
                symbolic x coords={L3b, L3i, L3b+3Lr}, 
                xtick=data,
                ymin=0, 
                ymax=60, 
                ytick = {0,10,20,30,40,50,60},
                enlarge x limits={abs=0.75cm},
                x tick label style={rotate=0, anchor=north},
            ]
            \addplot coordinates {(L3b, 31.42) (L3i, 53.98) (L3b+3Lr, 53.98)};  
            \addplot coordinates {(L3b, 29.15) (L3i, 48.72) (L3b+3Lr, 52.57)}; 
            \addplot coordinates {(L3b, 29.99) (L3i, 46.66) (L3b+3Lr, 53.09)}; 
            \addplot coordinates {(L3b, 30.51) (L3i, 43.92) (L3b+3Lr, 52.72)}; 
            \end{axis}
        \end{tikzpicture}
        \caption{Instruction Following}
    \label{fig:c_index_inst}
    \end{subfigure}
   ~ 
        \begin{subfigure}[t]{0.32\textwidth}
        \begin{tikzpicture}
            \begin{axis}[
                ybar, 
                ymajorgrids=true,
                bar width=6.5pt,
                width=180pt,
                height=6.5cm,
                legend style={font=\footnotesize,legend columns = -1},
                symbolic x coords={L3b, L3i, L3b+3Lr}, 
                xtick=data,
                ymin=70, 
                ymax=80, 
                enlarge x limits={abs=0.75cm},
                x tick label style={rotate=0, anchor=north},
            ]
            \addplot coordinates {(L3b, 75.85) (L3i, 74.79) (L3b+3Lr, 74.79)};  
            \addplot coordinates {(L3b, 76.29) (L3i, 75.18) (L3b+3Lr, 75.97)}; 
            \addplot coordinates {(L3b, 75.99) (L3i, 74.98) (L3b+3Lr, 76.12)}; 
            \addplot coordinates {(L3b, 75.85) (L3i, 74.98) (L3b+3Lr, 76)}; 
            \legend{org,100M,500M,1B} 
            \end{axis}
        \end{tikzpicture}
        \caption{Reasoning}
\label{fig:c_index_rea}
    \end{subfigure}
   ~ 
        \begin{subfigure}[t]{0.32\textwidth}
        \begin{tikzpicture}
            \begin{axis}[
                ybar, 
                ymajorgrids=true,
                bar width=6.5pt,
                width=180pt,
                height=6.5cm,
                symbolic x coords={L3b, L3i, L3b+3Lr}, 
                xtick=data,
                ymin=30, 
                ymax=60, 
                ytick  = {30,35,40,45,50,55,60},
                enlarge x limits={abs=0.75cm},
                x tick label style={rotate=0, anchor=north},
            ]
             \addplot coordinates {(L3b, 43.94) (L3i, 51.67) (L3b+3Lr, 51.67)};  
            \addplot coordinates {(L3b, 47.64) (L3i, 51.8) (L3b+3Lr, 56.13)}; 
            \addplot coordinates {(L3b, 47.29) (L3i, 51.22) (L3b+3Lr, 55.73)}; 
            \addplot coordinates {(L3b, 47.41) (L3i, 51.68) (L3b+3Lr, 56.17)}; 
            \end{axis}
        \end{tikzpicture}
        \caption{Truthfulness}
\label{fig:c_index_truth}
    \end{subfigure}
    \caption{Impact of continual pre-training on instruction following capability of LLaMa base and instruct models compared against residual technique. }
\label{fig:c_index}
\end{figure*}
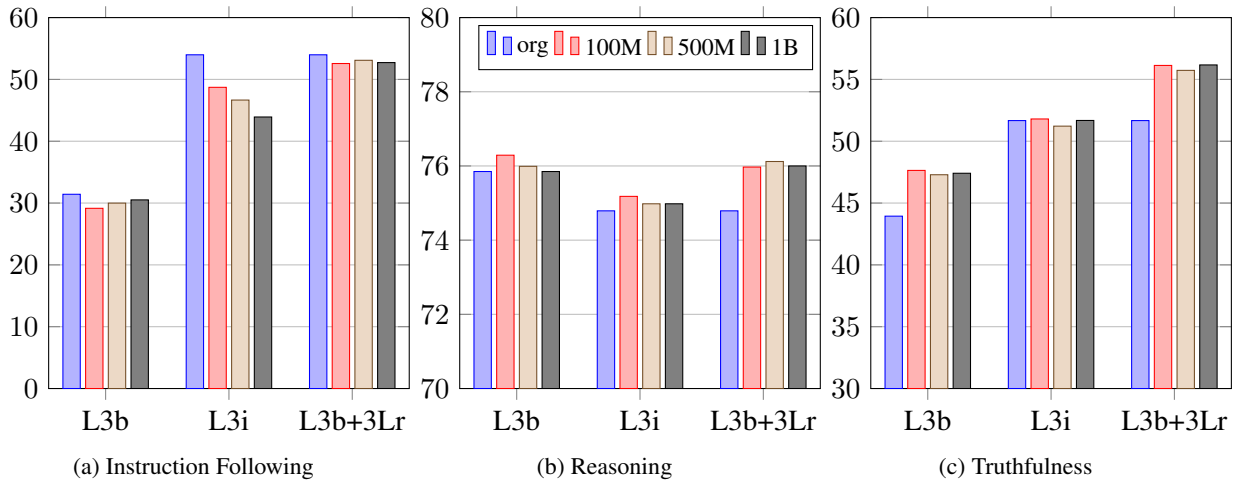

\subsection{Language Model Architectures}
\label{ssec:llm}
We used two distinct families of language models LLaMa \cite{dubey2024llama} and Qwen \cite{yang2024qwen2}. Specifically, we target the LLaMa 3, 3.1 family of models, and the Qwen 2, 2.5 family of models. Both LLaMa 3 and 3.1 are available in 8B, 70B parameters size with the exception that the 3.1 family also has a 405B parameters model. For all our LLaMa experiments we focused only on the 8B models because of the resource constraints. Similarly, both Qwen 2 and 2.5 come in 0.5B, 1.5B, and 7B parameter models with the exception that the 2.5 family also has 3B, 14B, 32B, and 72B parameter models. For all Qwen experiments, we choose 0.5B, 1.5B, and 7B models. Further, by design, we are required to choose a family of models for which both the base and the instruction-tuned variants of the same size exist.

\section{Results and Analysis}

In this section, we will examine the influence of continuous pre-training on the instructional capabilities of both the base model and its corresponding instruct models. Specifically, we will determine \textbf{if an additional round of fine-tuning is necessary after updating the Large Language Model (LLM) knowledge}. Subsequently, we will assess the efficacy of a proposed instruction residuals technique, which is simple yet effective, to \textbf{restore any lost instructional capabilities} without explicit instruction fine-tuning of the updated LLM.

\subsection{Impact of Continual Pretraining}
\label{ssec:icp}

In this section, we delve into the effects of continual pre-training on the LLaMa 3 8B models, specifically focusing on both the LLaMa 3 base (L3b) and the LLaMa 3 instruction-tuned (L3i) models. The pre-training process was carried out over a diverse set of new tokens, with quantities of 100M, 500M, and 1B, to comprehensively investigate the impact of the new token size on instructional capabilities (detailed in Section  \ref{ssec:pretrain}).

Figure \ref{fig:c_index} provides the summary of our findings. Notably, as the instruction model (L3i) encounters an increasing number of new tokens, its instructional capabilities deteriorate in Figure \ref{fig:c_index_inst}. For instance, with 100M new tokens, we observed a maximum drop of 5.7 points on the instruction following dataset and an average drop of 2.7 points across all tasks. This trend intensifies as we add more tokens to the continuous pre-training datasets. That is with 1B new tokens, resulting in a maximum drop of 10 points on the instruction following dataset and an average drop of 3 points overall. Hence, confirming that \textbf{the continual pre-training of instruction-tuned models with a large volume of new tokens leads to a significant loss of its instructional capabilities}.

In contrast, the base model (L3b) in Figure \ref{fig:c_index} demonstrates minimal quality degradation with the number of new tokens. While it is anticipated that base models to lack instructional capabilities, we do not observe significant catastrophic forgetting of any such capabilities with continual pre-training. Hence, \textbf{base models appear to be less susceptible to this effect, maintaining relatively stable performance despite an increase in new tokens}.

\begin{table*}[!t]
    \centering
\begin{adjustbox}{width=\linewidth}
    \begin{tabular}{ l l ||  l l l | l l l || l l l |  l l l}
    \toprule
\multicolumn{2}{c||}{LLM (Params) $\rightarrow$}  & \multicolumn{3}{c}{\textbf{LLaMa 3 (8B)}} & \multicolumn{3}{c||}{\textbf{LLaMa 3.1 (8B)}} & \multicolumn{3}{c}{\textbf{Qwen 2 (1.5B)}} & \multicolumn{3}{c}{\textbf{Qwen 2.5 (1.5B)}} \\ \cmidrule(lr){3-5} \cmidrule(lr){6-8} \cmidrule(lr){9-11} \cmidrule(lr){12-14}
\multirow{1}{*}{Benchmark}  & Metric  & L3b& L3i  &	+3.1Lr &	L3.1b &	L3.1i	& +3Lr & Q2b & Q2i & +2.5Qr & Q2.5b & Q2.5i& +2Qr\\ \midrule

 \multirow{4}{*}{\begin{tabular}[c]{@{}c@{}}IFEval\end{tabular}}  & ILL\_acc & 19.06 & \underline{53.36} & \textbf{57.19} & 15.59 &  \textbf{54.80} & \underline{48.32}  & 25.84 &  \textbf{29.02} & \underline{27.22} & \underline{28.15} &  \textbf{39.09} & 26.50  \\ 
 & ILS\_acc & 17.87 & \underline{47.84} & \textbf{51.92} & 14.63 &  \textbf{49.88} & \underline{43.76}  & 23.62 &  \textbf{26.74} & \underline{23.86} & \underline{25.72} &  \textbf{35.21} & 25.06 \\ 
 & PLL\_acc & 09.98 & \underline{41.77} & \textbf{43.81} & 07.76 &  \textbf{41.59} & \underline{34.38}  & 16.08 &  \textbf{17.93} & \underline{17.19} & 20.01 &  \textbf{25.67} & \underline{20.33} \\ 
 & PLS\_acc & 09.06 & \underline{35.30} & \textbf{37.52} & 07.02 &  \textbf{35.49} & \underline{29.39}  & 14.05 &  \textbf{15.71} & \underline{14.23} & 17.65 &  \textbf{22.04} & \underline{18.85} \\ \cmidrule{2-14}
MMLU & acc & 62.14 & \underline{63.83} & \textbf{66.02} & 63.40 &  \textbf{68.03} & \underline{64.89}  & 55.10 &  \textbf{55.80} & \underline{55.16} & 59.75 &  \textbf{59.75} & \underline{59.75}\\ 
MMLU-Pro & EM & 34.51 & \textbf{39.70} &\underline{ 38.53} & 35.32 & \textbf{40.97} & \underline{40.40}  & \underline{21.21}  & \textbf{21.70}  & 21.10 & \underline{27.13}  & \textbf{29.89}  & 27.00\\ \cmidrule{2-14} 
\multirow{2}{*}{GSM8K} & EM & 49.58 & \underline{75.06} & \textbf{75.13} & 50.11 &  \textbf{76.19} & \underline{73.84}  & 54.51 &  \textbf{58.30} & \underline{57.01} & \underline{57.39} & 56.86 & \textbf{58.76} \\ 
 & strict-EM & 49.20 & \underline{74.98} & \textbf{74.98} & 49.81 &  \textbf{75.36} & \underline{73.84} & 54.44 &  \textbf{57.39} & \underline{56.79} & \underline{57.16} & 53.37 &  \textbf{58.30} \\  \midrule \midrule
\multicolumn{2}{c||}{Sub-Average} & 31.43 &\underline{53.98} & \textbf{55.64} & 30.46 & \textbf{55.29} & \underline{51.10} & 33.11 & \textbf{35.32} & \underline{34.07} & 36.62 & \textbf{40.24} & \underline{36.82} \\ \midrule \midrule 
Winogrande & acc & \textbf{73.16} & 71.74 & \underline{72.77} &  \textbf{73.64} & \underline{73.40} & 73.01 &  \textbf{66.38} & 65.19 & \underline{65.82} & 63.22 &  \textbf{65.19} & \underline{63.22} \\ \cmidrule{2-14}
\multirow{2}{*}{Hellaswag} & acc & \textbf{60.12} & 57.70 & \underline{59.00} & \textbf{60.02} & \underline{59.10} & 58.25 & \underline{48.61} &  \textbf{49.30} & 48.58 & \underline{50.16} &  \textbf{50.94} & 50.08  \\  
 & acc\_n & \textbf{79.22} & 75.76 & \underline{78.88} & \underline{78.90} &  \textbf{79.19} & 76.68    & \underline{65.43} &  \textbf{66.07} & 65.42 & \underline{67.81} &  \textbf{68.34} & 67.76 \\ \cmidrule{2-14}
\multirow{2}{*}{ARC\_easy} & acc & 80.39 & \underline{81.52} & \textbf{81.57} & 81.40 & \underline{81.94} &  \textbf{82.91}& 66.25 &  \textbf{69.99} &\underline{66.37} & 75.42 &  \textbf{76.56} & \underline{75.67} \\ 
 & acc\_n & 77.78 & \textbf{79.63} & \underline{78.83} & \underline{81.14} & 79.76 &  \textbf{81.86}& \underline{60.86} &  \textbf{66.54} & 60.14 & 71.84 &  \textbf{76.26} & \underline{72.26}   \\ \cmidrule{2-14}
\multirow{2}{*}{Piqa} & acc & \textbf{79.54} & 78.56 & \underline{79.00} &  \textbf{80.09} & \underline{80.03} & 79.16 & 75.41 &  \textbf{76.17} & \underline{75.95} & 75.68 &  \textbf{76.39} & \underline{75.84} \\ 
 & acc\_n & \textbf{80.74} & 78.62 & \underline{80.14} & \underline{81.07} &  \textbf{81.12} & 79.22  & 75.41 &  \textbf{75.90} & \underline{75.79} & \underline{76.06} &  \textbf{76.12} & 75.79\\ \midrule \midrule
\multicolumn{2}{c||}{Sub-Average}  & \textbf{75.85} & 74.79 & \underline{75.74} & \textbf{76.61} & \underline{76.36} & 75.87 & 65.48 & \textbf{67.02} & \underline{65.48}& 68.60 & \textbf{69.97} & \underline{68.66} \\ \midrule \midrule
T\_mc2 & acc & 43.94 & \underline{51.67} & \textbf{52.73} & 45.17 &  \textbf{53.92} & \underline{52.20}  & \underline{45.95} & 43.36 &  \textbf{45.95} &  \textbf{46.64} & \underline{46.65} & 46.45\\ \midrule
\multicolumn{2}{c||}{Average}         & 51.64  & \underline{62.94}  & \textbf{64.25} & 51.57  &  \textbf{64.42}  & \underline{62.01}   & {48.07} &  \textbf{49.69} &\underline{48.54} & {51.24} &  \textbf{53.65} & \underline{51.35}\\ \bottomrule

    \end{tabular}
\end{adjustbox}
\caption{Instruction portability quality comparison with LLaMa 3 and 3.1 8B LLMs. Where the \textbf{BOLD} represents the best quality and the \underline{underline} shows the second-best score for that column block. }
\label{tab:results_portability}
\end{table*}

\subsection{Restore Instruction Capabilities}
\label{ssec:RIC}

As described in Section \ref{sec:ir} we restore the instructional capabilities by incorporating instructional residuals $\Theta^{v1}_{r}$ into a continuously pre-trained base model, where $\Theta^{v1}_{r}$ is calculated as the difference between the weights of the instruction tuned model $\theta^{d1v1}_{i}$ and the weights of the base model $\theta^{d1}_{b}$.

The results, as demonstrated in Table \ref{tab:results_pretrain} (last column block), reveal a significant improvement in model performance when the instruction residual technique is employed. Specifically, the residual-adjusted models (L3b + 3Lr) outperform the L3i models by an average of 4 absolute points across all tasks when pre-trained on 500M new tokens. This improvement is not only consistent but also increases with the number of new tokens, reaching a significant improvement of 5 absolute points when pre-trained on 1B new tokens.

Based on these findings, we assert that an \textbf{additional round of instruction fine-tuning is necessary after updating the instruction-tuned model's knowledge}. Furthermore, we posit that the \textbf{instructional residuals not only restore the instructional capabilities of the continuously pre-trained model} but also enhance these abilities on numerous tasks.

\subsection{Instruction Portability across LLM Families}
\label{ssec:ipl}
We summarize the impact of instruction residuals across different LLM families with varying model sizes in Table \ref{tab:results_portability}, where \emph{\{x\}b} represents the base model, \emph{\{x\}i} represents the corresponding instruction tuned model and \emph{+\{x\}r} represents the instruction residual adjusted continuously pre-trained base model. For example, \emph{+3.1Lr} means instruction residuals of LLaMa 3.1 family $\theta^{d1d2v1}_{L3.1i} - \theta^{d1d2}_{L3.1b}$  are added to the LLaMa 3 base model $\theta^{d1}_{L3b}$. Similarly, \emph{+2.5Qr} means instruction residuals of Qwen 2.5 family $\theta^{d1d2v1}_{Q2.5i} - \theta^{d1d2}_{Q2.5b}$  are added to the Qwen 2 base model $\theta^{d1}_{Q2b}$.

For the LLaMa family, we observe that the instruction residuals always improve the instruction-following capabilities of the base model. For both LLaMa 3 and 3.1 we observe a consistent gain over the base model for all the datasets. As noted in the LLaMa 3.1 technical report, it possesses high-quality instruction-following capabilities, Therefore,  LLaMa 3.1 instruction residuals are expected to carry better instruction-following capabilities than LLaMa 3's instruction residuals. As expected, instruction residuals of LLaMa 3.1 (\emph{3.1Lr}), when merged with the LLaMa 3 base model (\emph{L3b}), improve its instruction capabilities (64.25 Vs 51.64) better than its own instruction fine-tuned model (\emph{L3i}) by $~1$ absolute point. On the other hand, since LLaMa 3 is inferior in quality to LLaMa 3.1, its instruction residuals  (\emph{3Lr})  also possess low-quality instruction-following capabilities than instruction residuals of LLaMa 3.1 (\emph{3.1Lr}). Therefore, as expected, instruction residuals of LLaMa 3 (\emph{3Lr}) when merged with the LLaMa 3.1 base model (\emph{L3.1b}) improves its instruction capabilities (62.01 Vs 51.57). However, performs lower than its original instruction fine-tuned model (\emph{L3.1i}). One observation that stands out in this experiment is that the model with instruction residuals performs always better than the corresponding base models. Hence prove that \textbf{the instruction capabilities are portable across models of the same family LLMs}.

\subsection{Instruction Residual Applicability to Derived LLMs}

\label{ssec:ira}
In this section, we investigate the applicability of instruction residuals on the LLMs derived from the same ancestor. Specifically, we choose an LLM that is either continuously pre-trained or instruction-tuned on the LLaMa 3 base model. We find a plethora of LLMs derived from LLaMa 3 on HuggingFace \footnote{\url{https://huggingface.co/models?search=llama3}}. Among existing, we chose to experiment with 
\textit{cerebras/Llama3-DocChat-1.0-8B} as this model makes the LLaMa 3 base model specialized for a particular task. Specifically, this LLM adds a new document-based QA skill to the LLaMa 3 base model via instruction fine-tuning the base model with the ChatQA dataset. Table \ref{tab:result_docchat} summarizes the impact of instruction residuals on the instruction following capabilities of this LLM. It is evident that the instruction residual significantly improves the instruction-following capabilities of the DocChat LLM both with LLaMa 3 and 3.1 residuals. The impact is more pronounced with the LLaMa 3.1 residuals reaching a gain of $6$ absolute points in quality. Hence, it is evident that the \textbf{instruction residuals are portable across same ancestor models}.

\begin{table}[]
\begin{adjustbox}{width=0.9\linewidth}
\begin{tabular}{ll| lll}
\toprule
\textbf{Benchmark}         & \textbf{Metric}        & DocChat      & +3Lr      & +3.1Lr      \\ \midrule
\multirow{4}{*}{IFEval}    & ILL\_acc               & 38.25 & 49.64 & 56.71 \\
                           & ILS\_acc               & 34.65 & 46.04 & 53.12 \\
                           & PLL\_acc               & 24.95 & 37.34 & 44.36 \\
                           & PLS\_acc               & 20.89 & 32.90 & 39.74 \\ \cmidrule{2-5}
MMLU                       & acc                    & 62.96 & 62.31 & 65.35 \\
MMLU-Pro                   & EM                     & 36.36 & 39.66 & 39.36 \\ \cmidrule{2-5} 
\multirow{2}{*}{GSM8K}     & EM                     & 57.09 & 78.54 & 74.53 \\ 
                           & strict-EM                     & 56.94 & 77.48 & 69.90 \\ \midrule
Winogrande                 & acc                    & 74.27 & 71.27 & 73.32 \\ \cmidrule{2-5}
\multirow{2}{*}{Hellaswag} & acc                    & 61.68 & 57.57 & 59.51 \\
                           & acc\_n & 80.36 & 75.79 & 78.39 \\ \cmidrule{2-5}
\multirow{2}{*}{ARC\_easy} & acc                    & 82.11 & 81.02 & 80.22 \\
                           & acc\_n                 & 81.52 & 79.00 & 77.99 \\ \cmidrule{2-5}
\multirow{2}{*}{Piqa}      & acc                    & 80.47 & 78.13 & 78.78 \\
                           & acc\_n                 & 81.61 & 77.97 & 78.62 \\ \midrule
T\_mc2            & acc                    & 45.35 & 49.54 & 50.82 \\ \midrule
\multicolumn{2}{c|}{Average} & 57.47 & \underline{62.14} & \textbf{63.80}\\ \bottomrule
\end{tabular}
\end{adjustbox}
\caption{Applicability of instruction residual approach on publicly available \textit{Llama3-DocChat-1.0-8B} (DocChat) LLM that was built on top of LLaMa 3 base model. Here, +3Lr and +3.1Lr are the instruction residuals from LLaMa 3 and LLaMa 3.1, respectively integrated to the original DocChat LLM.}
\label{tab:result_docchat}
\end{table}

\section{Related Work}

\noindent\textbf{Continual Learning} is a well-established technique for updating existing machine learning models with the latest information and trends, allowing language models to adapt to new data while preserving the knowledge acquired during prior training \cite{caccia2020online,le2023bloom,ibrahim2024simple}. Several studies have applied continuous pre-training in large language models (LLMs) to acquire new skills, domains, and languages, and perform various tasks, as demonstrated in works such as \citet{yadav2023exploring,ma2023ecomgpt,yang2024pllama,gogoulou2023study}.

\citet{yang2024pllama} adopt a strategy of continuous pre-training followed by instruction fine-tuning on domain-specific data to learn new domains. However, the reasons behind why this approach is the most effective for acquiring both new knowledge and instruction-following capabilities in large language models (LLMs) have not been thoroughly explored.

\noindent\textbf{Model Merging}: Recent studies have demonstrated that specialized fine-tuned models can be merged to combine capabilities and generalize to new skills \cite{yu2024language,yang2024model}. Several techniques have been explored for merging the abilities of two or more models, including Task Arithmetic \cite{ilharco2022editing}, TIES \cite{yadav2024ties}, and Model Breadcrumbs \cite{davari2023model}. Following \citet{ilharco2022editing}, we employ Task Arithmetic in our work to extract the instruction residual. Although different model merging techniques could affect the transfer of instruction-following capabilities, understanding the specific impact of these techniques is beyond the scope of this study. 

\section{Conclusion}
 In conclusion, this study delves into the effects of continuous pre-training on base and instruction-tuned large language models (LLMs) and their instruction capabilities. The findings suggest that while continuous pre-training of instruction models may lead to catastrophic forgetting of instruction capabilities, a more efficient approach is to continuously pre-train the base model with new data, followed by instruction tuning. This method preserves both domain knowledge and instruction capabilities. Interestingly, the study also reveals that instruction capabilities are transferable across models from the same ancestor, eliminating the need for additional instruction tuning for a continuously pre-trained base model. We empirically demonstrated this analysis on the LLaMa 3 and LLaMa 3.1 family of base and instruction models.

\section*{Limitations} 
While our hypothesis is validated for models with 8 billion parameters, we observe a noticeable variation in performance when applied to smaller models, particularly those with around 1.5 billion parameters. Furthermore, the scalability of our proposed strategy for models smaller than 1.5 billion parameters remains uncertain. This presents an intriguing avenue for future research, where further exploration could investigate whether modifications or optimizations are needed to maintain the same level of effectiveness for these smaller models.

A critical challenge that emerges with the instruction residual method is the reliance on the availability of both the base language model and its instruction fine-tuned counterpart. The approach fundamentally depends on the residual differences between these two models to function effectively. In the absence of either the base model or the fine-tuned model, the instruction residual method cannot be employed. This limitation highlights a bottleneck in the methodology, especially when resources or computational constraints prevent the simultaneous availability of both models. Future work could explore potential ways to mitigate this dependency, perhaps by developing alternative techniques that either reduce the need for dual-model structures or enhance the portability of instruction-based fine-tuning across a wider range of model sizes.

\bibliography{custom}

\appendix

\section{Overview of Experimental Hardware}
\label{app:flops}

\subsection{GPU Specifications}
\begin{itemize}
\item GPU: NVIDIA A100 40GB SXM
\item GPU Memory: 40GB
\item FP16/BF16 Tensor Core: 312 TeraFLOPs
\item TF32: 156 TeraFLOPs
	
\end{itemize}

\subsection{FLOPs Requirements}

\noindent\textbf{Instruction Fine-tuning}
\begin{itemize}
	\item {Number of Parameters, \(N\)}: 8B.
        \item Number of tokens, \(tokens\)
\begin{align*}
\text{\(tokens\)} &: 25M samples \\
 &\approx 25M \times 8192 \\
 &= 204,800 \text{M tokens}
\end{align*}

	\item {Number of Epochs, \(E\)}: Fine-tuning generally requires fewer epochs; often 3 to 10 epochs are sufficient.
\end{itemize}

\noindent\textbf{Continued Pre-training}
The below calculations assume continuous pre-training with 100M Tokens.
\begin{itemize}
	\item {Number of Parameters, \(N\)}: 8B.
	\item {Number of tokens, \(tokens\)}: 100M tokens
	\item {Sequence Length, \(S\)}: 4096
	\item {Number of Epochs, \(E\)}: 5
\end{itemize}

\noindent\textbf{Estimate FLOPs per Tokens}

The FLOPs per training step depend on the number of operations performed per token per layer. Assuming each parameter needs about 6 floating-point operations (forward and backward):
\[
\text{FLOPs/token/parameter} \approx 6
\]

\noindent Given the structure of transformers with multiple layers and self-attention, let's simplify and assume each token requires 6 operations per parameter across all layers:
\vspace{-0.2cm}
\begin{align*}
\text{FLOPs/token} &= 6 \times N \\
\end{align*}


\subsection{A Comparison}
Here, we perform the comparison between LLaMa8B Instruction Tuning and 100M continous pre-training (CP) in terms of numbers of FLOPs. 

\begin{align*}
\textit{ratio} &= \frac{\text{Instruct}(6 \times 8 \times 10^9 \times tokens \times E)}{\text{CP}(6 \times 8 \times 10^9 \times tokens \times E)} \\
& = \frac{6 \times 10^9 \times 204800 Million \times 5}{6 \times 10^9 \times 100 Million \times 5} \\
& \approx 2048
\end{align*}

This calculation provides a rough estimate of the FLOPs required to continue pre-training the model on 100M tokens across 5 epoch and instruction fine-tuning FLOPs, estimates numbers from Llama3 Paper.

\subsection{MMLU Performance vs Compute Cost}

we demonstrate how our model maintained a good enough performance on the MMLU benchmark while optimizing for low computational costs. The results reflect efficient usage of available compute resources, particularly by leveraging hardware such as the NVIDIA A100.

\subsubsection{Comparison of MMLU Scores and Compute Costs}

\begin{figure}[h]
	\centering
	\includegraphics[width=0.50\textwidth]{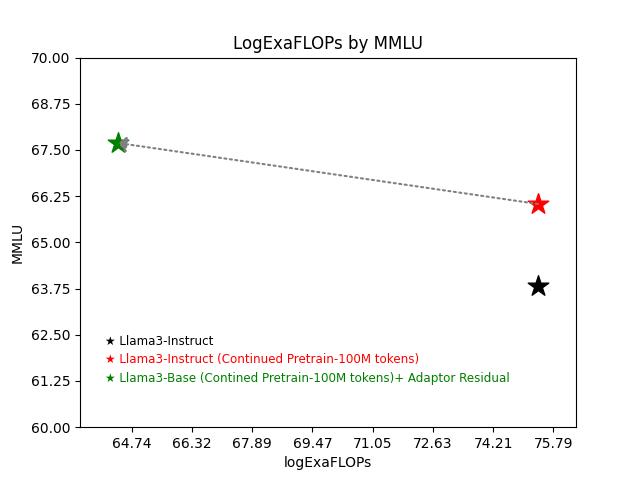}  
	\caption{Comparison of MMLU performance and compute costs across different models. Our model (marked in green) balances compute efficiency while maintaining competitive performance.}
	\label{fig:logflops_vs_mmlu}
\end{figure}

Our approach achieved high accuracy in various tasks under the MMLU benchmark, matching the performance of more compute-intensive models. Despite this, we successfully reduced the total compute cost by optimizing training and fine-tuning processes, as shown in Figure~\ref{fig:logflops_vs_mmlu}.

\end{document}